\documentclass[11pt]{article}
\usepackage{amssymb}
\usepackage{amsmath}

\usepackage[preprint]{acl}
\usepackage{booktabs}
\usepackage{subcaption}
\usepackage{times}
\usepackage{latexsym}
\usepackage{xcolor}
\usepackage{multirow}

\usepackage[T1]{fontenc}

\usepackage{makecell}

\usepackage{microtype}

\usepackage{inconsolata}

\usepackage{graphicx}

\title{When Does Language Matter? Multilingual Instructions Reveal Step-wise Language Sensitivity in Vision-Language-Action Models}

\author{Xuan Dong, Zhe Han, Tianhao Niu, Qingfu Zhu\thanks{Corresponding author.}, Wanxiang Che \\
  Harbin Institute of Technology \\
  \texttt{\{xd,zhan,thniu,qfzhu,car\}@ir.hit.edu.cn}}

\begin{document}
\maketitle
\begin{abstract}
Vision-Language-Action (VLA) models have shown strong performance in language-conditioned robotic manipulation, yet their robustness to linguistic variation remains poorly understood. In this work, we present the first systematic multilingual evaluation of VLA models by translating the LIBERO benchmark into ten languages, revealing severe performance degradation under non-English instructions, with success rates dropping by 30–50\%. Through fine-grained analysis of task executions, we find that language influence is highly non-uniform across steps: certain steps exhibit strong language dependence and dominate overall task failure, while others are largely language-agnostic. Based on this insight, we propose a step-wise inference-time intervention that aligns representations according to step language sensitivity, substantially improving performance under linguistic variation. Our results indicate that language robustness in VLA models is fundamentally a step-wise control problem, highlighting the importance of temporally structured analysis for reliable embodied agents.
\end{abstract}

\begin{figure}[t]
  \centering
  \includegraphics[width=\linewidth]{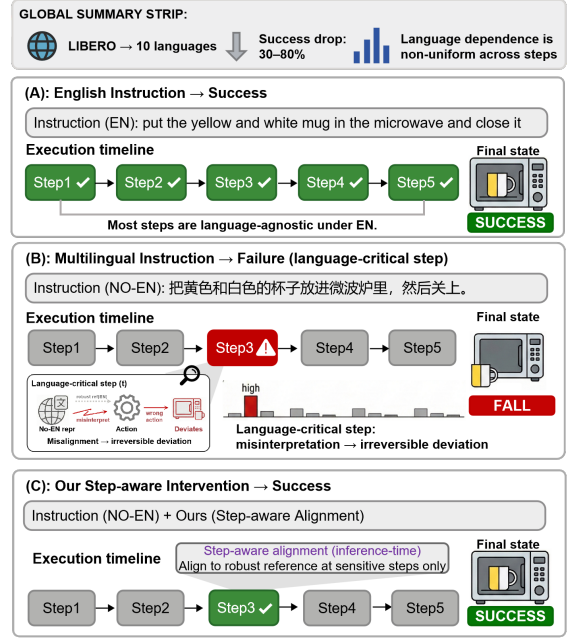}
  \caption{
  \textbf{Step-wise language sensitivity in VLA models.}
  Multilingual instructions lead to execution failures that concentrate at a subset of language-critical steps.
  By identifying these steps via step-wise analysis and selectively intervening during inference, we significantly improve robustness under linguistic variation.
  }
  \label{fig:overview}
\end{figure}

\section{Introduction}
Vision-Language-Action (VLA) models have emerged as a powerful paradigm for language-conditioned robotic manipulation, enabling agents to map visual observations and natural language instructions to long-horizon action sequences \cite{pmlr-v270-kim25c,brohan2023rt2visionlanguageactionmodelstransfer,black2024pi0visionlanguageactionflowmodel}. Recent advances have demonstrated impressive performance on standard manipulation benchmarks by scaling vision-language backbones and leveraging large-scale pretraining followed by task-specific fine-tuning \cite{kim2025finetuningvisionlanguageactionmodelsoptimizing,niu2024llarvavisionactioninstructiontuning}. However, these gains are often achieved at the expense of substantially increased training cost and exhibit limited robustness under distributional shifts, particularly in the language modality, where variations in instruction form can significantly affect downstream action execution \cite{guo2025robustnessvisionlanguageactionmodelmultimodal,liu2025evavlaevaluatingvisionlanguageactionmodels}.

To better understand language robustness in VLA models, we focus on multilingual instructions as a principled form of linguistic variation. While multilingual robustness has been extensively explored in language and vision-language models \cite{Viveiros2025TowerVisionUA,Manea2025MultilingualVM,maaz2024palopolyglotlargemultimodal}, its behavior in VLA models remains largely unexplored. This gap is critical because VLA models generate continuous action streams that directly affect the environment, causing language-induced errors to accumulate over time and irreversibly impact long-horizon execution. As a result, multilingual robustness in VLA models may manifest differently and warrants dedicated investigation.

To study this question, we translate the LIBERO \cite{liu2023liberobenchmarkingknowledgetransfer} benchmark into ten languages and conduct a systematic multilingual evaluation of VLA models. Our results reveal severe performance degradation under non-English instructions, with success rates dropping by 30–50\% compared to English. These findings indicate that, despite sharing architectural components with multilingual vision-language models, current VLA systems exhibit substantially weaker robustness to linguistic variation.

A natural approach to mitigating multilingual degradation is uniform alignment. For example, CLAIM~\cite{ye-etal-2025-claim} estimates an average cross-lingual representation shift between English and non-English inputs and applies this global correction at inference time. However, our analysis shows that deviations from English vary strongly across execution steps, rather than being evenly distributed. Averaging language shifts across all steps therefore weakens correction at steps where language differences are most pronounced, motivating step-wise intervention.

At the same time, alignment is not cost-free. Applying it indiscriminately can introduce noise at steps that are largely language-agnostic, as alignment inevitably perturbs representations that are already dominated by visual or proprioceptive signals rather than language. In a closed-loop setting, such perturbations may propagate through subsequent actions and alter future observations. This makes it important to restrict alignment to language-critical steps, rather than applying it uniformly across the execution horizon.

Motivated by the failure of step-agnostic alignment in VLA models, we analyze how language influences action prediction across execution steps and find that its effect is highly non-uniform over time.
Based on this insight, we propose a step-wise approach that explicitly respects the temporal structure of VLA execution. By applying alignment selectively at language-sensitive steps, our approach substantially improves multilingual robustness across languages and models, and further enables targeted, data-efficient training.
Figure~\ref{fig:overview} provides an overview of the step-wise language sensitivity and our intervention motivation.

Our contributions are summarized as follows:
\begin{itemize}
    \item We present the first systematic multilingual evaluation of VLA models, revealing severe performance degradation under linguistic variation.
    
    \item We conduct a step-wise analysis of VLA execution under linguistic variation, revealing that language influence is highly non-uniform across execution steps.

    \item We propose a step-wise approach for improving language robustness that respects step sensitivity, substantially improving performance under multilingual instructions and enabling data-efficient training.

\end{itemize}

\section{Related Work}

\subsection{Vision-Language-Action Models}

Vision-Language-Action (VLA) models map visual observations and natural language instructions directly to executable control actions, and have achieved strong performance through large-scale pretraining and task-specific fine-tuning, as demonstrated by RT-1 \cite{brohan2023rt1roboticstransformerrealworld}, RT-2 \cite{brohan2023rt2visionlanguageactionmodelstransfer}, Octo \cite{octomodelteam2024octoopensourcegeneralistrobot}, OpenVLA \cite{pmlr-v270-kim25c}, and $\pi_{0}$ \cite{black2024pi0visionlanguageactionflowmodel}. Accordingly, prior work largely focuses on model scaling, data efficiency, and generalization across tasks and environments \cite{zhou2025exploringlimitsvisionlanguageactionmanipulations,neary2025improvingpretrainedvisionlanguageactionpolicies}. In contrast, language inputs are typically assumed to be English, and robustness to linguistic variation, such as multilingual instructions, remains largely unexplored, leaving open how language shifts affect long-horizon action execution.

\subsection{Multilingual Vision-Language Models}

Multilingual robustness has been extensively studied in language and vision-language models, where performance gaps are commonly addressed through multilingual pretraining or fine-tuning \cite{chen2023palixscalingmultilingualvision,fan2025languagespecificlayermattersefficient}, as well as prompting strategies such as language-specific prompts or English chain-of-thought reasoning \cite{qin-etal-2023-cross,Zhao2025ACE}. More recently, inference-time intervention methods that align non-English representations toward English references have shown strong gains without additional training \cite{ye-etal-2025-claim,li2025unlockingmultilingualreasoningcapability}. However, these approaches do not readily transfer to Vision-Language-Action (VLA) models. Compared to large vision-language models, VLA systems emphasize mapping language and perception to low-level control actions rather than rich language understanding, making prompt-based or reasoning-heavy strategies less reliable. In addition, VLA models generate continuous action trajectories in a sequential, closed-loop manner, where global or representation-level alignment can introduce errors that propagate across execution steps. Consequently, treating multilingual robustness as a static alignment problem is often insufficient for embodied, long-horizon action execution.

These limitations motivate the need for a dedicated analysis of multilingual behavior in VLA models that accounts for their unique characteristics, including continuous action outputs and strong step-wise temporal dependencies. Our work addresses this gap by systematically analyzing multilingual robustness in VLA execution and proposing a step-wise approach that respects the temporal structure of action generation, without requiring additional training data.

\section{Methodology}


\subsection{Preliminary}
Vision-Language-Action (VLA) models formulate robotic manipulation as a multimodal control problem, where action prediction is conditioned on visual observations and natural language instructions.
At each execution step $t$, the agent receives an observation $\boldsymbol{o}_t$, typically consisting of image frames from onboard sensors and optional proprioceptive signals, together with a task-level natural language instruction $l$.
The policy then outputs a continuous action vector $\boldsymbol{a}_t \in \mathbb{R}^d$, where $d$ denotes the dimensionality of the action space, representing low-level control commands such as joint velocities or gripper actions.

Formally, a VLA model parameterized by $\theta$ implements a policy $\pi_\theta$ that maps observations and language instructions to actions:
\begin{equation}
\boldsymbol{a}_t = \pi_\theta(\boldsymbol{o}_t, l),
\end{equation}
where the same instruction $l$ is provided throughout the entire execution, while the observation $\boldsymbol{o}_t$ evolves over time.
Internally, the model encodes visual observations and language instructions into sequences of visual tokens and language tokens, which are processed by a Transformer backbone \cite{NIPS2017_3f5ee243} to produce multimodal representations for action prediction.

Recent VLA architectures cast action generation as a conditional sequence modeling problem, where actions are produced through autoregressive token prediction or flow matching. This formulation enables VLAs to perform long-horizon, language-conditioned manipulation, but also introduces complex interactions between linguistic inputs and action generation across execution steps.

\subsection{Step-wise Language Sensitivity Analysis}
\label{sec:stepwise_sensitivity}
To characterize how language information influences action prediction across execution steps, we conduct a step-wise analysis of language sensitivity at the representation levels.
First, we quantify \emph{where} multilingual deviations occur by comparing hidden representations induced by an English instruction and its non-English counterparts~\cite{ye-etal-2025-claim}. For each step $t$, we extract a hidden representation $\boldsymbol{h}_t$ from a fixed layer and measure its deviation from the English-conditioned representation, revealing a strongly step-dependent pattern.

\begin{figure*}[t]
    \centering
    \includegraphics[width=\textwidth]{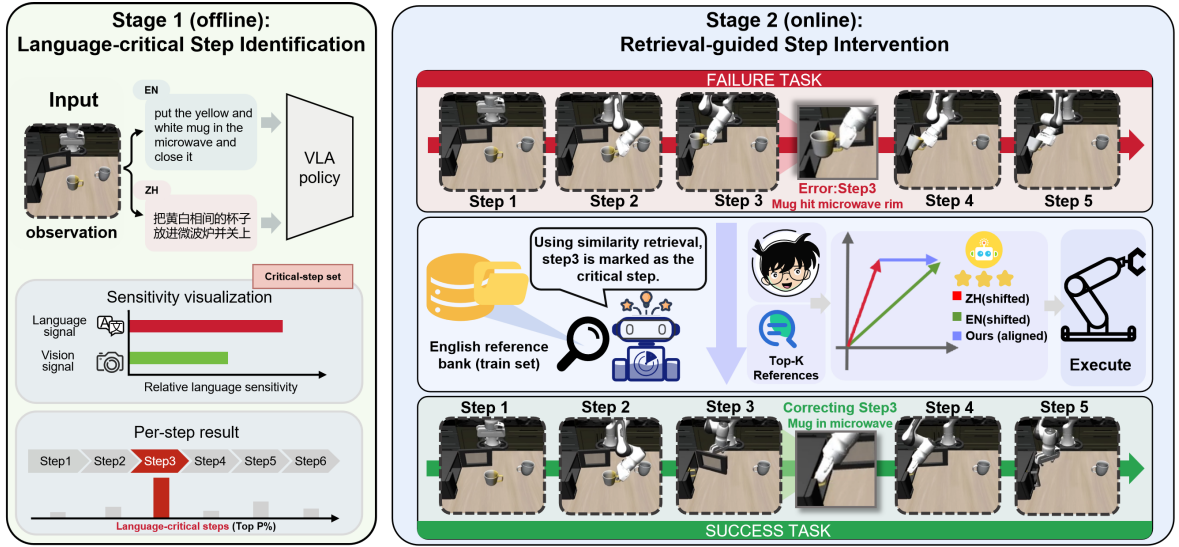} 
    \vspace{-2mm}
    \caption{\textbf{Overview of our step-wise pipeline.}
    \textbf{Stage 1 (offline)} identifies language-critical steps via step-wise sensitivity analysis.
    \textbf{Stage 2 (online)} retrieves the nearest English reference and checks whether the current step is language-critical.
    \textbf{Stage 3 (online)} performs step-wise alignment by retrieving top-$K$ English neighbors and updating the current action representation, improving multilingual rollouts without retraining.}
    \label{fig:method_overview}
    \vspace{-2mm}
\end{figure*}

However, representation deviations are mainly diagnostic: they explain \emph{where} differences appear, but are less actionable for identifying \emph{why} particular steps are language-sensitive and for supporting inference-time step selection without English references.
We therefore estimate intrinsic language reliance using gradients on English executions.
Specifically, we compute the gradients of the predicted action $\boldsymbol{a}_t$ with respect to language tokens and visual tokens, and derive scalar sensitivity scores by averaging over tokens and dimensions.
Using English as a pivot yields an efficient, language-agnostic criterion for locating language-sensitive steps, which generalizes well across languages (Section~\ref{sec:gradient_analysis}).

For each step $t$, let $L$ and $V$ denote the sets of language and visual tokens.
We define the average gradient magnitudes as

\begin{equation}
\begin{aligned}
g_t^{\text{lang}} &= \frac{1}{|L|} \sum_{x \in L}
\left\lVert \frac{\partial a_t}{\partial x} \right\rVert, \\
g_t^{\text{vis}} &= \frac{1}{|V|} \sum_{x \in V}
\left\lVert \frac{\partial a_t}{\partial x} \right\rVert .
\end{aligned}
\end{equation}

We then quantify step-wise language sensitivity using the \emph{text--image gradient ratio}
\begin{equation}
r_t = \frac{g_t^{\text{lang}}}{g_t^{\text{vis}} + \epsilon},
\end{equation}
where $\epsilon$ is a small constant added for numerical stability.
A larger value of $r_t$ indicates that action prediction at step $t$ relies more heavily on language inputs relative to visual observations.

We find that steps with larger gradient ratios tend to coincide with steps exhibiting larger deviations between non-English and English action representations at the same step.
This correspondence suggests that gradient-based language sensitivity provides a meaningful signal for identifying execution steps where linguistic variation is most likely to affect action generation.
Based on this signal, we categorize steps with high $r_t$ as language-sensitive and those with low $r_t$ as language-agnostic, which serves as the basis for our subsequent step-wise intervention strategies.

\subsection{Step-wise Alignment During Inference}

Based on the step-wise language sensitivity analysis, we introduce an inference-time alignment strategy that selectively intervenes at language-critical execution steps.
Our method operates on a hidden representation of the VLA model and applies alignment independently at each step.
Figure~\ref{fig:method_overview} summarizes the overall step-wise pipeline, including offline sensitivity identification and online selective alignment.

\paragraph{Reference Set Construction.}
We first construct a reference set from a subset of English training trajectories.
For each training sample $i$ and execution step $t$, we extract the hidden representation at a fixed intermediate layer of the VLA model, denoted as $\tilde{h}_t^{(i)} \in \mathbb{R}^{d_h}$, where $d_h$ denotes the dimensionality of the hidden representation.
Each reference representation is indexed by its execution step and associated with a pre-computed language sensitivity score obtained from the step-wise gradient analysis.
We denote the resulting reference set as
\[
\mathcal{R} = \{ \tilde{\boldsymbol{h}}_t^{(i)} \}.
\]

\paragraph{Step-wise Retrieval.}
At inference time, given a non-English instruction and the current observation at step $t$, the VLA model produces a hidden representation $\boldsymbol{h}_t$ at the same layer.
We retrieve the $K$ nearest reference representations from $\mathcal{R}$ according to cosine similarity:
\begin{equation}
\mathcal{N}_t =
\operatorname{Top}\text{-}K_{\tilde{\boldsymbol{h}} \in \mathcal{R}}
\left(
\frac{\boldsymbol{h}_t^\top \tilde{\boldsymbol{h}}}{\|\boldsymbol{h}_t\| \, \|\tilde{\boldsymbol{h}}\|}
\right).
\end{equation}

\paragraph{Language-critical Step Identification.}
Let $\mathcal{C} \subset \mathcal{R}$ denote the subset of reference representations whose corresponding steps rank within the top $p\%$ according to the language sensitivity metric.
To determine whether alignment should be applied at step $t$, we examine the retrieved neighborhood $\mathcal{N}_t$ and define a gating indicator
\begin{equation}
\mathbb{I}_t =
\mathbb{1}
\left(
\frac{|\mathcal{N}_t \cap \mathcal{C}|}{|\mathcal{N}_t|}
\ge \tau
\right),
\end{equation}
where $\tau \in (0,1)$ specifies the minimum fraction of language-critical neighbors required to trigger intervention.
If $\mathbb{I}_t = 0$, no alignment is performed at step $t$.

\paragraph{Similarity-weighted Alignment.}
When $\mathbb{I}_t = 1$, we compute a reference hidden representation by aggregating the retrieved neighbors using similarity-based weights:
\begin{equation}
\bar{\boldsymbol{h}}_t
=
\sum_{\tilde{\boldsymbol{h}}^{(i)} \in \mathcal{N}_t}
w_i \, \tilde{\boldsymbol{h}}^{(i)}.
\label{eq:ref_agg}
\end{equation}

\vspace{-1.5em}

\begin{equation}
w_i
=
\frac{\exp\!\left(\beta \, \cos(\boldsymbol{h}_t, \tilde{\boldsymbol{h}}^{(i)})\right)}
{\sum_{\tilde{\boldsymbol{h}}^{(j)} \in \mathcal{N}_t}
\exp\!\left(\beta \, \cos(\boldsymbol{h}_t, \tilde{\boldsymbol{h}}^{(j)})\right)} .
\label{eq:sim_weight}
\end{equation}

where $\beta$ controls the sharpness of the weighting distribution.
The aligned hidden representation is then obtained via a step-wise update:
\begin{equation}
\boldsymbol{h}_t^{\text{aligned}} =
\boldsymbol{h}_t + \alpha \, \mathbb{I}_t \, (\bar{\boldsymbol{h}}_t - \boldsymbol{h}_t),
\end{equation}
where $\alpha$ controls the strength of the alignment.
The updated representation $\boldsymbol{h}_t^{\text{aligned}}$ is subsequently passed to the remaining layers of the VLA model to produce the final action for execution.
All operations are performed at inference time without additional training or optimization.

\section{Experiment}

\subsection{Experimental Settings}

\paragraph{Models.}
We conduct experiments on two Vision-Language-Action models, OpenVLA-OFT \cite{kim2025finetuningvisionlanguageactionmodelsoptimizing} and $\pi_{0.5}$ \cite{intelligence2025pi05visionlanguageactionmodelopenworld}. These models represent different VLA architectures and training setups, allowing us to evaluate the behavior of step-wise interventions across model variants.

\paragraph{Languages.}
To evaluate robustness under linguistic variation, we consider ten languages in total, including English and nine non-English languages: Chinese, French, Japanese, Korean, Spanish, Portuguese, Arabic, Thai, and Vietnamese.
For each task, the original English instructions are translated into the target languages using Google Translate. Translation quality was evaluated via human checking and back translation. See Appendix~\ref{app:translation_quality} for more details.

\paragraph{Benchmark.}
All experiments are conducted on the LIBERO benchmark, a standardized evaluation suite for language-conditioned robotic manipulation.
LIBERO comprises four long-horizon task suites—\emph{Spatial}, \emph{Object}, \emph{Goal}, and \emph{Long}—designed to test generalization across spatial layouts, object identities, task goals, and temporal compositions.
We refer readers to Appendix~\ref{app:libero_overview} for a detailed description of the benchmark and task suites.

\paragraph{Evaluation Metric.}
We report success rate as the primary evaluation metric, defined as the percentage of episodes in which the task is completed successfully. All reported results are averaged over multiple evaluation episodes.

\paragraph{Baselines.}
We compare our method against three representative baselines for handling linguistic variation.
\textbf{(i) Multilingual (No Intervention)} executes non-English instructions directly, reflecting the default behavior of current VLA models.
\textbf{(ii) English chain-of-thought (EN-CoT)}~\cite{liu-etal-2025-translation} prepends an English reasoning prompt (e.g., “think step by step in English”) to encourage English-centric intermediate reasoning without modifying model internals.
\textbf{(iii) Average Alignment}~\cite{ye-etal-2025-claim} applies a uniform correction by estimating an average representation shift between English and non-English instructions from training data and applying it at all execution steps during inference.

\paragraph{Experimental Details.}
All hyperparameters are fixed across experiments unless stated otherwise.
Due to space constraints, the main text reports results averaged over LIBERO task suites.
Full per-task multilingual results and implementation details are provided in Appendix~\ref{app:full_results_tables} and Appendix~\ref{app:details}, respectively.
We additionally evaluate inference latency and find that step-wise intervention introduces negligible additional computational cost and detailed results are reported in Appendix~\ref{app:latency}.

\subsection{Multilingual Evaluation on LIBERO}

We first evaluate the multilingual performance of VLA models on the LIBERO benchmark without any intervention. For each task, we report success rates under ten languages, including English and nine non-English translations of the original instructions, in order to quantify the extent of performance degradation induced by linguistic variation.

\begin{table*}[t]
\centering
\small
\setlength{\tabcolsep}{4pt}
\renewcommand{\arraystretch}{0.9}

\begin{subtable}[t]{\textwidth}
\centering
\caption{OpenVLA-OFT on LIBERO (success rate, \%).}
\label{tab:multilingual_libero_openvla}
\begin{tabular*}{\textwidth}{@{\extracolsep{\fill}}lcccccccccc@{}}
\toprule
\textbf{Task} & \textbf{EN} & \textbf{ZH} & \textbf{JA} & \textbf{KO} & \textbf{FR} & \textbf{ES} & \textbf{AR} & \textbf{TH} & \textbf{PT} & \textbf{VI} \\
\midrule
Spatial &
\textbf{97.6} &
55.8{\scriptsize\textcolor{red}{-41.8}} &
56.8{\scriptsize\textcolor{red}{-40.8}} &
49.2{\scriptsize\textcolor{red}{-48.4}} &
64.0{\scriptsize\textcolor{red}{-33.6}} &
63.8{\scriptsize\textcolor{red}{-33.8}} &
52.8{\scriptsize\textcolor{red}{-44.8}} &
53.0{\scriptsize\textcolor{red}{-44.6}} &
62.6{\scriptsize\textcolor{red}{-35.0}} &
59.2{\scriptsize\textcolor{red}{-38.4}} \\

Object &
\textbf{98.4} &
76.0{\scriptsize\textcolor{red}{-22.4}} &
92.0{\scriptsize\textcolor{red}{-6.4}} &
78.6{\scriptsize\textcolor{red}{-19.8}} &
97.6{\scriptsize\textcolor{red}{-0.8}} &
97.8{\scriptsize\textcolor{red}{-0.6}} &
73.2{\scriptsize\textcolor{red}{-25.2}} &
78.6{\scriptsize\textcolor{red}{-19.8}} &
96.6{\scriptsize\textcolor{red}{-1.8}} &
91.8{\scriptsize\textcolor{red}{-6.6}} \\

Goal &
\textbf{97.9} &
9.8{\scriptsize\textcolor{red}{-88.1}} &
11.4{\scriptsize\textcolor{red}{-86.5}} &
11.6{\scriptsize\textcolor{red}{-86.3}} &
15.2{\scriptsize\textcolor{red}{-82.7}} &
11.2{\scriptsize\textcolor{red}{-86.7}} &
6.4{\scriptsize\textcolor{red}{-91.5}} &
10.6{\scriptsize\textcolor{red}{-87.3}} &
12.0{\scriptsize\textcolor{red}{-85.9}} &
10.8{\scriptsize\textcolor{red}{-87.1}} \\

Long &
\textbf{94.5} &
65.0{\scriptsize\textcolor{red}{-29.5}} &
77.2{\scriptsize\textcolor{red}{-17.3}} &
69.6{\scriptsize\textcolor{red}{-24.9}} &
84.4{\scriptsize\textcolor{red}{-10.1}} &
83.4{\scriptsize\textcolor{red}{-11.1}} &
70.8{\scriptsize\textcolor{red}{-23.7}} &
70.6{\scriptsize\textcolor{red}{-23.9}} &
83.0{\scriptsize\textcolor{red}{-11.5}} &
76.4{\scriptsize\textcolor{red}{-18.1}} \\

Average &
\textbf{97.1} &
51.7{\scriptsize\textcolor{red}{-45.4}} &
59.4{\scriptsize\textcolor{red}{-37.7}} &
52.3{\scriptsize\textcolor{red}{-44.8}} &
65.3{\scriptsize\textcolor{red}{-31.8}} &
64.1{\scriptsize\textcolor{red}{-33.0}} &
50.8{\scriptsize\textcolor{red}{-46.3}} &
53.2{\scriptsize\textcolor{red}{-43.9}} &
63.6{\scriptsize\textcolor{red}{-33.5}} &
59.6{\scriptsize\textcolor{red}{-37.5}} \\

\bottomrule
\end{tabular*}
\end{subtable}

\vspace{6pt}

\begin{subtable}[t]{\textwidth}
\centering
\caption{$\pi_{0.5}$ on LIBERO (success rate, \%).}
\label{tab:multilingual_libero_pi05}
\begin{tabular*}{\textwidth}{@{\extracolsep{\fill}}lcccccccccc@{}}
\toprule
\textbf{Task} & \textbf{EN} & \textbf{ZH} & \textbf{JA} & \textbf{KO} & \textbf{FR} & \textbf{ES} & \textbf{AR} & \textbf{TH} & \textbf{PT} & \textbf{VI} \\
\midrule
Spatial &
\textbf{97.8} &
67.8{\scriptsize\textcolor{red}{-30.0}} &
70.0{\scriptsize\textcolor{red}{-27.8}} &
69.8{\scriptsize\textcolor{red}{-28.0}} &
72.8{\scriptsize\textcolor{red}{-25.0}} &
73.6{\scriptsize\textcolor{red}{-24.2}} &
68.0{\scriptsize\textcolor{red}{-29.8}} &
68.6{\scriptsize\textcolor{red}{-29.2}} &
75.4{\scriptsize\textcolor{red}{-22.4}} &
69.6{\scriptsize\textcolor{red}{-28.2}} \\

Object &
\textbf{98.8} &
69.6{\scriptsize\textcolor{red}{-29.2}} &
71.2{\scriptsize\textcolor{red}{-27.6}} &
67.8{\scriptsize\textcolor{red}{-31.0}} &
80.8{\scriptsize\textcolor{red}{-18.0}} &
77.8{\scriptsize\textcolor{red}{-21.0}} &
68.4{\scriptsize\textcolor{red}{-30.4}} &
68.8{\scriptsize\textcolor{red}{-30.0}} &
75.6{\scriptsize\textcolor{red}{-23.2}} &
72.6{\scriptsize\textcolor{red}{-26.2}} \\

Goal &
\textbf{97.6} &
16.8{\scriptsize\textcolor{red}{-80.8}} &
12.8{\scriptsize\textcolor{red}{-84.8}} &
11.4{\scriptsize\textcolor{red}{-86.2}} &
13.6{\scriptsize\textcolor{red}{-84.0}} &
14.2{\scriptsize\textcolor{red}{-83.4}} &
12.8{\scriptsize\textcolor{red}{-84.8}} &
14.8{\scriptsize\textcolor{red}{-82.8}} &
13.4{\scriptsize\textcolor{red}{-84.2}} &
13.6{\scriptsize\textcolor{red}{-84.0}} \\

Long &
\textbf{93.4} &
72.4{\scriptsize\textcolor{red}{-21.0}} &
71.8{\scriptsize\textcolor{red}{-21.6}} &
72.6{\scriptsize\textcolor{red}{-20.8}} &
77.6{\scriptsize\textcolor{red}{-15.8}} &
70.8{\scriptsize\textcolor{red}{-22.6}} &
73.4{\scriptsize\textcolor{red}{-20.0}} &
71.8{\scriptsize\textcolor{red}{-21.6}} &
73.0{\scriptsize\textcolor{red}{-20.4}} &
73.2{\scriptsize\textcolor{red}{-20.2}} \\

Average &
\textbf{96.9} &
56.7{\scriptsize\textcolor{red}{-40.2}} &
56.5{\scriptsize\textcolor{red}{-40.4}} &
55.4{\scriptsize\textcolor{red}{-41.5}} &
61.2{\scriptsize\textcolor{red}{-35.7}} &
59.1{\scriptsize\textcolor{red}{-37.8}} &
55.7{\scriptsize\textcolor{red}{-41.2}} &
56.0{\scriptsize\textcolor{red}{-40.9}} &
59.4{\scriptsize\textcolor{red}{-37.5}} &
57.3{\scriptsize\textcolor{red}{-39.6}} \\

\bottomrule
\end{tabular*}
\end{subtable}

\caption{Multilingual evaluation on LIBERO across ten languages. We report success rates (\%) for four task suites and the average. Red values indicate absolute performance drops relative to English.}
\label{tab:multilingual_libero_both}
\end{table*}

Table~\ref{tab:multilingual_libero_both} summarizes the results across four LIBERO task suites, together with the average success rate over all tasks, for both OpenVLA-OFT and $\pi_{0.5}$. The results show substantial performance variation across languages, highlighting the challenges faced by current VLA models when generalizing beyond English instructions.


\subsection{Non-uniform Language Influence Across Steps}
\paragraph{Non-uniform language influence.}

We begin by examining how non-English instructions deviate from English during long-horizon action execution.
For each task and language, we measure the step-wise difference between hidden representations generated under non-English instructions and those generated under English, using cosine distance in the hidden embedding space.
Importantly, English is not included in the visualization, such that all values reflect deviation relative to English.

\begin{figure}[t]
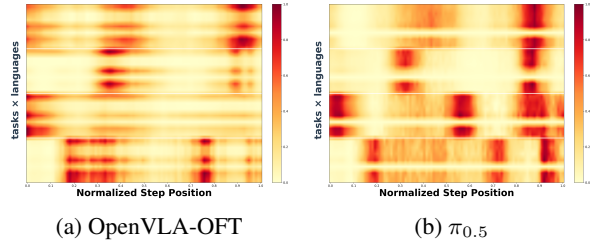

    \centering
    \begin{subfigure}[t]{0.48\columnwidth}
        \centering
        \includegraphics[width=\textwidth]{figures/combined_all_tasks_cosine_heatmap.pdf}
        \caption{OpenVLA-OFT}
        \label{fig:stepwise_deviation_openvla}
    \end{subfigure}
    \hfill
    \begin{subfigure}[t]{0.48\columnwidth}
        \centering
        \includegraphics[width=\textwidth]{figures/pi05_stepwise_heatmap.pdf}
        \caption{$\pi_{0.5}$}
        \label{fig:stepwise_deviation_pi05}
    \end{subfigure}

    \vspace{-2mm}
    \caption{
    Step-wise deviation from English across tasks and models.
    Each heatmap shows the cosine distance between hidden representations produced under non-English instructions and those produced under English at each execution step.
    The x-axis denotes normalized execution steps, while the y-axis indexes language–task combinations, grouped by LIBERO task suites with different non-English languages within each group.
    Darker colors indicate larger deviations from English.
    }
    \label{fig:stepwise_deviation}
\end{figure}

Figure~\ref{fig:stepwise_deviation} visualizes step-wise deviations from English for four LIBERO task suites across ten languages.
For each model, the horizontal axis denotes normalized execution steps.
The vertical axis is organized into four contiguous blocks corresponding to the Spatial, Object, Goal, and Long task suites, respectively, with each block containing rows for the non-English languages.
Across all task suites and both models (OpenVLA-OFT and $\pi_{0.5}$), we observe a clear non-uniform temporal pattern: deviations from English are not evenly distributed over the execution horizon, but instead concentrate at a subset of execution steps, forming distinct temporal hotspots shared across languages within the same task.

\begin{figure*}[t]
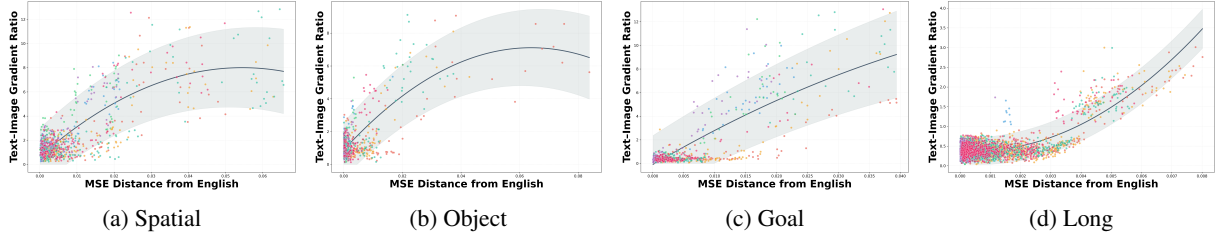

    \centering
    \begin{subfigure}[t]{0.24\textwidth}
        \centering
        \includegraphics[width=\textwidth]{figures/libero_spatial_mse_ratio_scatter.pdf}
        \caption{Spatial}
        \label{fig:grad_vs_dev_spatial}
    \end{subfigure}
    \hfill
    \begin{subfigure}[t]{0.24\textwidth}
        \centering
        \includegraphics[width=\textwidth]{figures/libero_object_mse_ratio_scatter.pdf}
        \caption{Object}
        \label{fig:grad_vs_dev_object}
    \end{subfigure}
    \hfill
    \begin{subfigure}[t]{0.24\textwidth}
        \centering
        \includegraphics[width=\textwidth]{figures/libero_goal_mse_ratio_scatter.pdf}
        \caption{Goal}
        \label{fig:grad_vs_dev_goal}
    \end{subfigure}
    \hfill
    \begin{subfigure}[t]{0.24\textwidth}
        \centering
        \includegraphics[width=\textwidth]{figures/libero_10_mse_ratio_scatter.pdf}
        \caption{Long}
        \label{fig:grad_vs_dev_long}
    \end{subfigure}

    \vspace{-2mm}
    \caption{
    Each point corresponds to an execution step (aggregated over rollouts), with the x-axis showing the step-wise deviation from English (MSE in representation space) and the y-axis showing the text–image gradient ratio.
    Points corresponding to different non-English languages are shown in different colors.
    Across tasks, steps with larger deviations from English consistently exhibit higher gradient ratios, supporting gradient-based sensitivity as an indicator of language-critical steps.
    }
    \label{fig:grad_vs_dev}
\end{figure*}

\begin{table*}[t]
\centering
\small
\setlength{\tabcolsep}{4pt}
\renewcommand{\arraystretch}{0.9}

\begin{tabular*}{\textwidth}{@{\extracolsep{\fill}}llcccccccccc@{}}
\toprule
\textbf{Model} & \textbf{Method} & \textbf{EN} & \textbf{ZH} & \textbf{JA} & \textbf{KO} & \textbf{FR} & \textbf{ES} & \textbf{AR} & \textbf{TH} & \textbf{PT} & \textbf{VI} \\
\midrule

\multirow{4}{*}{\makecell[c]{OpenVLA-OFT}}
& Baseline
& 97.1 & 51.7 & 59.4 & 52.3 & 65.3 & 64.1 & 50.8 & 53.2 & 63.6 & 59.6 \\
& EN-CoT
& -- & 56.8 & 54.8 & 52.0 & 62.4 & 63.4 & 51.1 & 52.8 & 62.3 & 55.1 \\
& Average shift
& -- & 60.8 & 58.9 & 54.7 & 65.5 & 64.5 & 53.8 & 54.5 & 64.2 & 57.1 \\
& \textbf{Step-wise shift (ours)}
& -- & \textbf{68.8} & \textbf{68.8} & \textbf{66.4} & \textbf{70.9} & \textbf{69.8} & \textbf{64.5} & \textbf{62.6} & \textbf{70.7} & \textbf{68.0} \\
\midrule

\multirow{4}{*}{\makecell[c]{$\pi_{0.5}$}}
& Baseline
& 96.9 & 56.7 & 56.5 & 55.4 & 61.2 & 59.1 & 55.7 & 56.0 & 59.4 & 57.3 \\
& EN-CoT
& -- & 55.2 & 55.7 & 55.7 & 59.9 & 57.9 & 56.4 & 56.5 & 57.2 & 56.3 \\
& Average shift
& -- & 54.1 & 53.9 & 52.9 & 57.2 & 56.2 & 53.0 & 52.6 & 55.6 & 54.2 \\
& \textbf{Step-wise shift (ours)}
& -- & \textbf{80.3} & \textbf{80.9} & \textbf{79.8} & \textbf{82.0} & \textbf{81.1} & \textbf{81.8} & \textbf{79.9} & \textbf{82.1} & \textbf{81.9} \\
\bottomrule
\end{tabular*}

\caption{Average success rates (\%) on LIBERO comparing baseline, EN-CoT, average, and step-wise alignment across ten languages and two VLA models.}

\label{tab:avg_vs_stepwise}
\end{table*}

These observations highlight a fundamental temporal structure in how language affects VLA execution.
Rather than inducing a global shift in behavior, multilingual variation manifests in a step-wise manner, with a subset of steps accounting for a disproportionate amount of deviation from English.
This motivates a closer examination of which steps are most sensitive to language input and how such sensitivity relates to downstream task failure.

\paragraph{Gradient-based language sensitivity.}
\label{sec:gradient_analysis}
Having established that deviations from English are temporally localized, we next ask whether such deviations are associated with steps where action prediction relies more on language than vision.
For each execution step, we compute the \emph{text--image gradient ratio} (Section~\ref{sec:stepwise_sensitivity}) and examine its relationship with the step-wise deviation from English.

\begin{figure}[t]
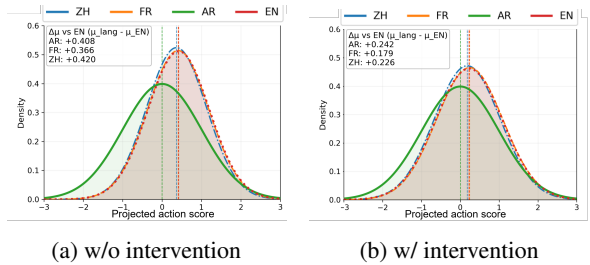

    \centering
    \begin{subfigure}[t]{0.48\linewidth}
        \centering
        \includegraphics[width=\linewidth]{figures/hidden_dist_standard.pdf}
        \caption{w/o intervention}
        \label{fig:hidden_dist_standard}
    \end{subfigure}
    \hfill
    \begin{subfigure}[t]{0.48\linewidth}
        \centering
        \includegraphics[width=\linewidth]{figures/hidden_dist_intervened.pdf}
        \caption{w/ intervention}
        \label{fig:hidden_dist_intervened}
    \end{subfigure}

    \caption{
    Hidden representation distributions projected onto a fixed English reference direction.
    }
    \label{fig:hidden_dist}
\end{figure}

Figure~\ref{fig:grad_vs_dev} shows that steps with larger deviations from English tend to exhibit higher text--image gradient ratios across all four LIBERO task suites.
This strong step-level association supports using gradient-based language sensitivity as a principled indicator of language-critical decision points.
Motivated by this finding, we use the gradient ratio to identify language-sensitive steps and to guide inference-time alignment, aiming to reduce non-English drift from the English reference.

\begin{table*}[t]
\centering
\small
\setlength{\tabcolsep}{4pt}
\renewcommand{\arraystretch}{0.95}

\begin{tabular*}{\textwidth}{@{\extracolsep{\fill}}llccccccccc@{}}
\toprule
\textbf{Model} & \textbf{Method} &
\textbf{ZH} & \textbf{JA} & \textbf{KO} & \textbf{FR} & \textbf{ES} &
\textbf{AR} & \textbf{TH} & \textbf{PT} & \textbf{VI} \\
\midrule

\multirow{4}{*}{\textbf{OpenVLA-OFT}} 
& Baseline
& 51.7 & 59.4 & 52.3 & 65.3 & 64.1 & 50.8 & 53.2 & 63.6 & 59.6 \\

& Random-step shift
& 65.9 & 62.2 & 62.5 & 66.7 & 67.7 & 54.1 & 53.1 & 68.1 & 62.5 \\

& All-step shift
& \underline{66.7} & \underline{67.6} & \underline{66.1} & \underline{69.9} & \underline{69.6}
& \underline{63.7} & \textbf{64.0} & \underline{69.7} & \underline{66.6} \\
& \textbf{Critical-step shift}
& \textbf{68.8} & \textbf{68.8} & \textbf{66.4} & \textbf{70.9} & \textbf{69.8} &
\textbf{64.5} & \underline{62.6} & \textbf{70.7} & \textbf{68.0} \\

\midrule

\multirow{4}{*}{$\boldsymbol{\pi_{0.5}}$}
& Baseline
& 56.7 & 56.5 & 55.4 & 61.2 & 59.1 & 55.7 & 56.0 & 59.4 & 57.3 \\

& Random-step shift
& 65.8 & 64.9 & 63.5 & 67.7 & 66.5 & 63.8 & 65.3 & 66.2 & 65.1 \\

& All-step shift
& \underline{79.6} & \underline{79.6} & \underline{78.9} & \underline{81.9} & \underline{80.5} & \underline{80.4} & \textbf{81.8} & \underline{80.3} & \underline{81.7} \\

& \textbf{Critical-step shift}
& \textbf{80.3} & \textbf{80.9} & \textbf{79.8} & \textbf{82.0} & \textbf{81.1} & 
  \textbf{81.8} & \underline{79.9} & \textbf{82.1} & \textbf{81.9} \\

\bottomrule
\end{tabular*}

\caption{
Which steps matter for step-wise intervention.
We report average success rates (\%) on LIBERO across non-English languages.
}
\label{tab:which_steps_matter}
\end{table*}

\begin{table}[t]
\centering
\small
\setlength{\tabcolsep}{4pt}
\renewcommand{\arraystretch}{1.05}

\begin{tabular}{lccccc}
\toprule
\textbf{Strategy} &
\textbf{Spatial} &
\textbf{Object} &
\textbf{Goal} &
\textbf{Long} &
\textbf{Avg.} \\
\midrule
Baseline    & 55.8 & 76.0 & 9.8 & 65.0 & 51.7 \\
Full         & 94.0 & 97.8 & 76.4 & 74.0 & 86.8 \\
Random-step   & 88.8 & 90.8 & 0.0 & 0.0 & 44.9 \\
\textbf{Critical-step} & 92.6 & 96.0 & 72.0 & 81.2 & 85.5 \\
\bottomrule
\end{tabular}

\caption{
Data-efficient fine-tuning on OpenVLA-OFT with Chinese instructions.
We report success rates (\%) on four LIBERO task suites and their average.
}
\label{tab:data_efficient_training}
\end{table}

\subsection{Main Results}
\paragraph{Gradient-guided Step-wise Alignment Improves Multilingual Robustness.}
Table~\ref{tab:avg_vs_stepwise} summarizes average success rates on LIBERO across ten languages for OpenVLA-OFT and $\pi_{0.5}$.
Across both models, gradient-guided step-wise alignment consistently improves multilingual performance over the no-intervention baseline, yielding substantial gains for most non-English languages.
This improvement is robust across task suites and does not rely on language-specific tuning.
In comparison, EN-CoT yields limited and inconsistent gains, underscoring the difficulty of transferring prompt-based reasoning strategies to action-generating policies.
Average alignment, which applies a uniform correction across all execution steps, leads to unstable effects and can even degrade performance.
These results indicate that treating language influence as temporally uniform is insufficient for VLA execution.

By contrast, step-wise alignment guided by gradient-based language sensitivity selectively corrects language-critical decision points.
To further validate this effect, we visualize the distribution of hidden representations before and after intervention.
As shown in Figure~\ref{fig:hidden_dist}, step-wise alignment consistently shifts non-English representations toward the English reference, confirming that targeted intervention at language-critical steps effectively reduces cross-lingual deviation.

\paragraph{Data-Efficient Training with Step-wise Selection.}
Finally, we examine whether step-wise language sensitivity can be exploited to improve training efficiency.
Rather than fine-tuning uniformly over all execution steps, we use the identified language-critical steps to selectively guide parameter updates.
Specifically, we select the language-sensitive steps according to the gradient-based analysis and restrict fine-tuning to these steps.

We compare three training strategies:
(i) full fine-tuning using all training data and all execution steps,
(ii) fine-tuning only on the selected language-critical steps (50\% of steps), and
(iii) fine-tuning on a randomly selected 50\% subset of steps for a fair comparison.

Table~\ref{tab:data_efficient_training} reports the results.
Fine-tuning on language-critical steps consistently outperforms random step selection, indicating that these steps capture meaningful language–action dependencies.
Although full fine-tuning achieves the best performance by leveraging all training data at a higher computational cost, step-wise selective fine-tuning recovers a large fraction of the gains with substantially fewer updates.
These results indicate that step-wise analysis is not only effective for inference-time intervention, but also enables more data-efficient training.

\section{Analysis}
\subsection{Do Language-Critical Steps Drive Performance?}
We further examine whether the effectiveness of step-wise intervention depends on which execution steps are aligned. Specifically, we compare three strategies under the same alignment mechanism:
(i) intervening only at the identified language-critical steps,
(ii) intervening at a randomly selected subset of steps with the same cardinality, and
(iii) intervening uniformly at all execution steps.
This comparison isolates the effect of step selection from the alignment operation itself.

Table~\ref{tab:which_steps_matter} summarizes the results.
Intervening at language-critical steps consistently yields higher success rates than random step intervention across languages and models, indicating that the identified steps are not interchangeable with arbitrary subsets.
Intervening at all steps further improves performance over random selection, but does not consistently surpass critical-step intervention, suggesting that unnecessary alignment at language-agnostic steps may introduce noise. This analysis shows that selectively intervening at language-critical steps is more effective than random or all-step intervention.

\subsection{Do Language-Critical Steps Correspond to Specific Action Primitives?}
We further ask whether language-critical failures are associated with particular manipulation primitives. To answer this, we analyze 100 multilingual failure cases randomly sampled from unsuccessful rollouts and attribute each case to the first action primitive at which the execution becomes irrecoverably incorrect. We group failures into six primitive categories: navigation to the target, grasp, place, toggle/rotate, open/close, and push. As shown in Table~\ref{tab:primitive_failure_breakdown}, failures concentrate most strongly on \textit{Navigate} (53\%), followed by \textit{Grasp} (18\%), while the remaining primitives account for substantially smaller shares.

This pattern suggests that multilingual degradation is driven primarily by errors in instruction-to-visual grounding, especially when the model must identify the correct object, receptacle, or target location from the instruction before acting. More than half of all sampled failures occur at the navigation stage, indicating that language-critical steps are closely tied to semantic target selection rather than being uniformly distributed across low-level motor primitives. By comparison, failures in \textit{Place}, \textit{Open/Close}, \textit{Toggle/Rotate}, and \textit{Push} are less frequent, further supporting the view that the dominant source of multilingual failure is incorrect grounding of instruction semantics in the visual scene.

\begin{table}[t]
\centering
\small
\setlength{\tabcolsep}{6pt}
\renewcommand{\arraystretch}{1.05}
\begin{tabular}{@{}l p{3.6cm} c@{}}
\toprule
\textbf{Primitive} & \textbf{Description} & \textbf{Rate (\%)} \\
\midrule
Navigate & Move to the intended target location/receptacle & 53 \\
Grasp & Pick up the intended target object & 18 \\
Place & Place the object to the correct receptacle/target & 10 \\
Toggle/Rotate & Toggle/rotate a control (e.g., knobs/switches) & 7 \\
Open/Close & Open/close drawers/cabinets & 9 \\
Push & Push the intended object to the desired location & 3 \\
\bottomrule
\end{tabular}
\caption{Primitive-level breakdown of 100 sampled multilingual failure cases. Each case is assigned to the first action primitive where execution becomes irrecoverably incorrect.}
\label{tab:primitive_failure_breakdown}
\end{table}

\subsection{How Do Language-Critical Steps Lead to Failure?}
To further illustrate how language-critical steps lead to downstream failures, we provide qualitative visualizations of execution trajectories highlighting representative failure cases in Appendix~\ref{app:qualitative}.

\section{Conclusion}

In this work, we conduct the first systematic multilingual evaluation of Vision-Language-Action models and show that their performance degrades substantially under linguistic variation. Through step-wise analysis, we find that language influence on action prediction is highly non-uniform over time, with certain language-critical steps disproportionately driving task failure. Motivated by this observation, we demonstrate that step-agnostic interventions are ineffective, while step-wise intervention significantly improves multilingual robustness at inference time. We further show that step-wise language sensitivity enables data-efficient training by guiding selective fine-tuning. Overall, our results highlight that language robustness in embodied agents is fundamentally a step-wise control problem and call for temporally structured analysis and intervention in future VLA systems.

\section*{Limitations}

Our study has several limitations. First, all experiments are conducted in simulation environments. While this setting enables controlled and reproducible analysis of multilingual robustness, it remains an open question how the observed step-wise language sensitivity and the proposed interventions transfer to real-world robotic systems, where additional sources of perception noise, actuation uncertainty, and embodiment constraints are present. Second, our findings are grounded in the current generation of Vision-Language-Action models, whose generalization capabilities are still limited. As VLA models evolve toward more general and robust embodiments, it will be important to revisit whether step-wise language sensitivity persists in the same form and how our analysis and intervention strategies should adapt to future model architectures and training paradigms.

\section*{Acknowledgments}

We gratefully acknowledge the support of the National Natural Science Foundation of China (NSFC) via grant 62236004 and 62476073.


\bibliography{custom}
\clearpage
\appendix
\section{Multilingual LIBERO Benchmark}
\label{app:multilingual_libero}

\subsection{LIBERO Benchmark Overview}
\label{app:libero_overview}

LIBERO is a standardized benchmark for lifelong robotic manipulation designed to evaluate generalization under controlled distribution shifts.
It provides a unified manipulation interface while systematically varying task structure, object identity, spatial layout, and temporal composition, making it well suited for analyzing long-horizon action execution.

LIBERO consists of four task suites.
\textbf{LIBERO-Spatial} evaluates spatial generalization by varying object layouts while keeping object identities fixed.
\textbf{LIBERO-Object} focuses on object-level generalization by varying object identities under fixed layouts.
\textbf{LIBERO-Goal} varies task goals while keeping objects and layouts fixed, requiring goal-conditioned reasoning.
\textbf{LIBERO-Long} combines variations in objects, layouts, and goals, resulting in long-horizon, multi-stage manipulation tasks.

Each task is specified by a natural language instruction and executed as a sequence of low-level control actions.
In this work, LIBERO serves as a controlled testbed for studying how linguistic variation affects step-wise action execution in Vision-Language-Action models.
Table~\ref{tab:libero_translation_example} shows an example LIBERO instruction and its translations into several target languages.

\subsection{Translation Quality Evaluation}
\label{app:translation_quality}

To verify the quality of the multilingual instructions used in our experiments, we evaluate translation consistency via back-translation.
Specifically, we translate all non-English LIBERO instructions back into English using Google Translate and compare them with the original English instructions.

We assess translation quality by measuring whether VLA model predictions conditioned on back-translated English instructions are consistent with those conditioned on the original English instructions.
High consistency indicates that the translated instructions preserve task semantics and do not introduce unintended ambiguity or semantic drift.

Table~\ref{tab:translation_quality} reports the average prediction consistency across languages, demonstrating that the translated instructions maintain high semantic fidelity and are suitable for evaluating multilingual robustness.

\begin{table}[t]
\centering
\small
\setlength{\tabcolsep}{6pt}
\renewcommand{\arraystretch}{1.05}
\begin{tabular}{lp{0.65\linewidth}}
\toprule
\textbf{Lang.} & \textbf{Instruction} \\
\midrule
EN & Put both the cream cheese box and the butter in the basket. \\
FR & Mettez la boîte de fromage à la crème et le beurre dans le panier. \\
ES & Coloque la caja de queso crema y la mantequilla en la cesta. \\
PT & Coloque a caixa de cream cheese e a manteiga na cesta. \\
\bottomrule
\end{tabular}
\caption{Example LIBERO instruction and its multilingual translations.}
\label{tab:libero_translation_example}
\end{table}

\begin{table}[t]
\centering
\small
\setlength{\tabcolsep}{2.7pt}
\renewcommand{\arraystretch}{1.05}

\makebox[\linewidth][c]{%
\begin{tabular}{lccccccccc}
\toprule
\textbf{Lang.} & ZH & JA & KO & FR & ES & AR & TH & PT & VI \\
\midrule
\textbf{Consistency (\%)} & 99 & 99 & 98 & 99 & 99 & 98 & 99 & 99 & 98 \\
\bottomrule
\end{tabular}
}

\caption{Back-translation consistency (\%) between original English instructions and back-translated English instructions across languages.}
\label{tab:translation_quality}
\end{table}

\section{Additional Experimental Details}
\label{app:details}

We provide additional implementation details for the step-wise alignment method used in our experiments.
Unless otherwise specified, all hyperparameters are fixed across tasks and languages.

\paragraph{Alignment strength $\alpha$.}
The scalar $\alpha$ controls the strength of the alignment applied to the hidden representation at each selected execution step.
We use different values of $\alpha$ for different VLA models to account for differences in representation scale and action sensitivity.

For \textbf{OpenVLA-OFT}, we set $\alpha = 1.0$, which consistently yields strong improvements across languages without destabilizing execution.
For $\boldsymbol{\pi_{0.5}}$, we use a slightly smaller value $\alpha = 0.8$, as larger alignment strengths were observed to occasionally introduce over-correction in long-horizon trajectories.
These values are selected based on validation performance and held fixed for all reported experiments.

\paragraph{Number of retrieved references $K$.}
At each execution step, we retrieve the top-$K$ nearest English reference representations based on cosine similarity.
We set $K=5$ in all experiments, which provides a good balance between robustness and stability.
Smaller values of $K$ lead to higher variance in the aligned representation, while larger values offer diminishing returns and increase computational overhead.
We find that performance is relatively insensitive to $K$ within the range $[3, 8]$, and thus adopt $K=5$ as a default choice.

\paragraph{Critical-step threshold.}
Language-critical steps are identified based on the step-wise language sensitivity metric derived from gradient analysis.
We select the top $50\%$ of execution steps with the highest language sensitivity scores as the critical set.
This threshold is used consistently for both inference-time alignment and data-efficient fine-tuning experiments.
Empirically, we observe that performance is stable across a reasonable range of thresholds (e.g., $40\%$--$60\%$), and thus fix the threshold at $50\%$ for simplicity.

\paragraph{Model-specific implementation.}
All alignment operations are performed at a fixed intermediate Transformer layer for each model.
For OpenVLA-OFT and $\pi_{0.5}$, we select the same relative layer depth (approximately the middle of the backbone ~\cite{ye-etal-2025-claim}), which we find to provide a good trade-off between language sensitivity and execution stability.
No model parameters are modified during inference-time alignment.

\section{Qualitative Analysis}
\label{app:qualitative}

To further illustrate how language-critical steps lead to downstream failures, we provide qualitative visualizations of execution trajectories under non-English instructions.
We present representative qualitative examples from the LIBERO-Object and LIBERO-Goal task suites.

\begin{figure*}[t]
    \centering

    \begin{subfigure}[t]{\textwidth}
        \centering
        \includegraphics[width=\textwidth]{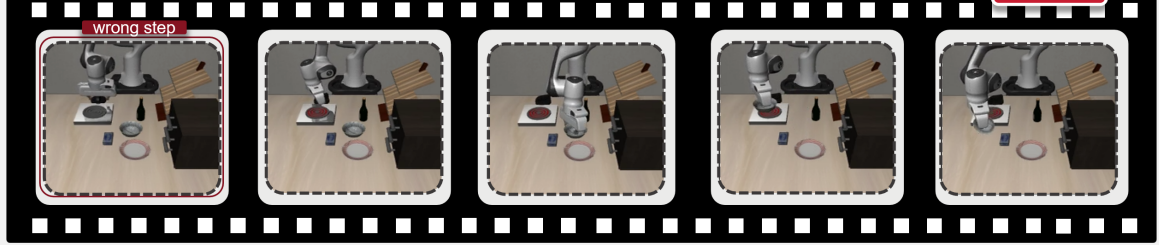}
        \caption{LIBERO-Object}
        \label{fig:qual_object}
    \end{subfigure}

    \vspace{4pt}

    \begin{subfigure}[t]{\textwidth}
        \centering
        \includegraphics[width=\textwidth]{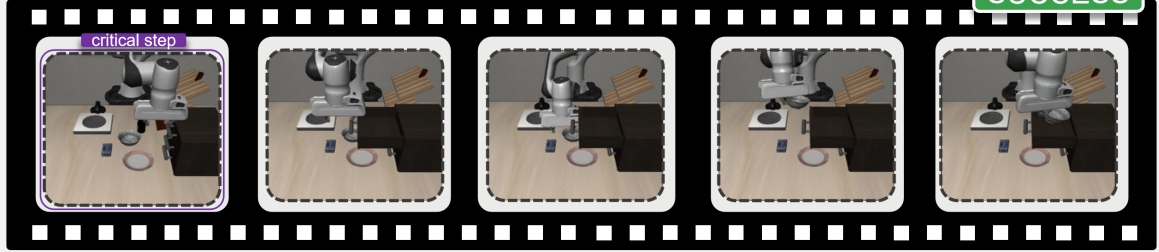}
        \caption{LIBERO-Goal}
        \label{fig:qual_goal}
    \end{subfigure}

    \caption{
    Qualitative examples illustrating how language-critical steps lead to task failure.
    For each example, we compare a non-English rollout without intervention (top) and a rollout with step-wise alignment applied at the identified language-critical step (bottom).
    Without intervention, errors at language-critical steps (highlighted in red) lead to irreversible failures, such as incorrect object or receptacle selection.
    Applying alignment at these steps corrects the action decision and enables successful task completion.
    }
    \label{fig:qualitative_examples}
\end{figure*}

Each example compares a non-English rollout without intervention to a rollout where step-wise alignment is applied at the identified language-critical step.
Across task suites, these examples show that multilingual failures typically originate at specific decision-critical steps rather than accumulating gradually over time.
Errors at such steps often irreversibly alter the environment state, leaving little opportunity for recovery in later execution.
By intervening precisely at these language-critical steps, our method directly targets the root cause of multilingual failure in long-horizon VLA execution.

\section{Additional Experimental Results}
\label{app:full_results}

\subsection{Inference Latency Analysis}
\label{app:latency}

We evaluate the computational cost introduced by step-wise intervention by measuring the average execution time per task on an NVIDIA A100 GPU.
For each LIBERO task suite, we report the mean latency over all non-English languages, computed across complete rollouts under identical settings.

Table~\ref{tab:latency} shows that step-wise intervention introduces only negligible runtime overhead.
Across all task suites, execution latency with alignment remains comparable to the baseline without intervention, with an average increase of approximately 1.3\%.
This indicates that the proposed method improves multilingual robustness without materially sacrificing runtime efficiency.

\begin{table}[t]
\centering
\small
\setlength{\tabcolsep}{8pt}
\renewcommand{\arraystretch}{1.05}
\begin{tabular}{lcc}
\toprule
\textbf{Task Suite} & \textbf{Baseline (s) $\downarrow$} & \textbf{Step-wise (s) $\downarrow$} \\
\midrule
Spatial & 13.58 & 13.72 \\
Object  & 15.08 & 15.18 \\
Goal    & 21.94 & 20.98 \\
Long    & 24.50 & 26.22 \\
\midrule
Average & 18.78 & 19.03 \scriptsize{(+1.3\%)} \\
\bottomrule
\end{tabular}
\caption{Average execution latency per task (seconds), averaged over all non-English languages and measured on an NVIDIA A100 GPU.
Lower is better. Relative change is computed with respect to the baseline.}
\label{tab:latency}
\end{table}

\subsection{Full Per-Task Multilingual Results}
\label{app:full_results_tables}

Due to space constraints, the main paper reports only average success rates across task suites when comparing intervention methods.
In this appendix, we provide the full per-task results for all languages to offer a more complete view of model behavior under multilingual instructions.

Tables~\ref{tab:full_results_compact} report detailed success rates on the four LIBERO task suites (\emph{Spatial}, \emph{Object}, \emph{Goal}, and \emph{Long}) for each non-English language.
Results include the baseline performance, English chain-of-thought prompting, average intervention, step-wise intervention based on all steps and gradient-identified language-critical steps.

Consistent with the trends observed in the main paper, step-wise intervention guided by gradient-based language sensitivity improves performance across most tasks and languages, while average or all-step interventions yield limited or unstable gains.
Random step selection is consistently inferior to gradient-based step selection, further supporting the importance of identifying language-critical execution steps.

\begin{table*}[t]
\centering
\small
\setlength{\tabcolsep}{2.5pt}
\renewcommand{\arraystretch}{0.95}
\caption{Full per-task multilingual results on LIBERO for OpenVLA-OFT and $\pi_{0.5}$ (success rate, \%).}
\label{tab:full_results_compact}
\begin{tabular*}{\textwidth}{@{\extracolsep{\fill}}llccccc|ccccc@{}}
\toprule
& & \multicolumn{5}{c}{\textbf{OpenVLA-OFT}} & \multicolumn{5}{c}{$\boldsymbol{\pi_{0.5}}$} \\
\cmidrule(lr){3-7} \cmidrule(lr){8-12}
\textbf{Lang.} & \textbf{Method}
& Spatial & Object & Goal & Long & Average
& Spatial & Object & Goal & Long & Average \\
\midrule

\multirow{6}{*}{ZH}
& Baseline        & 55.8 & 76.0 & 9.8  & 65.0 & 51.65 & 67.8 & 69.6 & 16.8 & 72.4 & 56.65 \\
& EN-CoT          & 55.6  & 87.4  & 11.8  & 72.2  & 56.75  & 69.6 & 68.2 & 14.4 & 68.6 & 55.20 \\
& Average shift   & 60.4 & 95.8 & 6.8  & 80.2 & 60.80 & 67.8 & 69.4 & 10.4 & 68.6 & 54.05 \\
& All-step shift  & 77.2 & 93.0 & 11.0 & 85.0 & 66.55 & 87.6 & 87.2 & 57.4 & 86.0 & 79.6\\
& Random-step     & 73.0 & 91.4 & 11.4 & 87.6 & 65.85 & 74.6 & 79.6 & 29.8 & 79.4 & 65.85 \\
& Critical-step   & 77.8 & 93.4 & 13.2 & 90.8 & 68.80 & 89.2 & 88.6 & 53.2 & 90.0 & 80.3 \\
\midrule

\multirow{6}{*}{JA}
& Baseline        & 56.8 & 92.0 & 11.4 & 77.2 & 59.35 & 70.0 & 71.2 & 12.8 & 71.8 & 56.45 \\
& EN-CoT          & 55.4  & 82.0  & 10.2  & 71.6  & 54.80  & 71.0 & 68.6 & 12.2 & 71.0 & 55.70 \\
& Average shift   & 58.8 & 90.0 & 6.2  & 80.4 & 58.85 & 66.0 & 70.0 & 8.4  & 71.0 & 53.85 \\
& All-step shift  & 75.6 & 96.4 & 11.8 & 86.4 & 67.55  & 87.2 & 89.4 & 57.6 & 84.0 & 79.55 \\
& Random-step     & 60.2 & 92.0 & 11.6 & 84.8 & 62.15 & 75.2 & 79.8 & 25.0 & 79.6 & 64.90 \\
& Critical-step   & 76.2 & 96.8 & 11.8 & 90.2 & 68.75 & 88.4 & 90.0 & 55.0 & 90.0 & 80.85 \\
\midrule

\multirow{6}{*}{KO}
& Baseline        & 49.2 & 78.6 & 11.6 & 69.6 & 52.25 & 69.8 & 67.8 & 11.4 & 72.6 & 55.40 \\
& EN-CoT          & 48.2  & 81.4  & 10.2  & 68.0  & 51.95  & 71.2 & 65.0 & 12.6 & 74.0 & 55.70 \\
& Average shift   & 56.2 & 81.8 & 5.4  & 75.2 & 54.65 & 69.4 & 65.2 & 7.6  & 69.2 & 52.85 \\
& All-step shift  & 71.6 & 95.6 & 13.4 & 83.8 & 66.10 & 89.8 & 89.4 & 52.4 & 84.0 & 78.90 \\
& Random-step     & 64.4 & 90.2 & 11.6 & 83.8 & 62.50 & 78.4 & 77.2 & 20.2 & 78.2 & 63.50 \\
& Critical-step   & 71.8 & 94.4 & 13.4 & 86.0 & 66.40 & 89.2 & 89.6 & 52.4 & 88.0 & 79.80 \\
\midrule

\multirow{6}{*}{FR}
& Baseline        & 64.0 & 97.6 & 15.2 & 84.4 & 65.30 & 72.8 & 80.8 & 13.6 & 77.6 & 61.20 \\
& EN-CoT          & 60.6  & 96.6  & 16.4  & 80.4  & 63.50 & 73.4 & 76.2 & 11.4 & 78.6 & 59.90 \\
& Average shift   & 65.2 & 96.0 & 16.2 & 84.4 & 65.45 & 71.0 & 75.0 & 7.6  & 75.2 & 57.20 \\
& All-step shift  & 82.6 & 96.6 & 15.0 & 85.4 & 69.90 & 90.8 & 90.6 & 56.2 & 90.0 & 81.90 \\
& Random-step     & 76.6 & 96.0 & 10.0 & 84.0 & 66.65 & 80.2 & 84.4 & 22.8 & 83.4 & 67.70 \\
& Critical-step   & 84.8 & 97.4 & 16.0 & 85.4 & 70.90 & 90.0 & 89.0 & 59.0 & 90.0 & 82.00 \\
\midrule

\multirow{6}{*}{ES}
& Baseline        & 63.8 & 97.8 & 11.2 & 83.4 & 64.05 & 73.6 & 77.8 & 14.2 & 70.8 & 59.10 \\
& EN-CoT          & 62.4  & 95.4  & 12.6  & 79.2  & 62.40  & 72.4 & 73.6 & 13.4 & 72.0 & 57.85 \\
& Average shift   & 65.2 & 97.0 & 11.8 & 84.0 & 64.50 & 70.8 & 76.4 & 10.6 & 66.8 & 56.15 \\
& All-step shift  & 85.2 & 97.2 & 11.4 & 84.6 & 69.60 & 88.0 & 90.8 & 61.0 & 82.0 & 80.45 \\
& Random-step     & 76.6 & 97.0 & 10.0 & 87.2 & 67.70 & 79.4 & 81.4 & 26.4 & 79.0 & 66.55 \\
& Critical-step   & 83.2 & 97.4 & 12.0 & 86.4 & 69.75 & 89.0 & 89.4 & 62.0 & 84.0 & 81.10 \\
\midrule

\multirow{6}{*}{AR}
& Baseline        & 52.8 & 73.2 & 6.4  & 70.8 & 50.80 & 68.0 & 68.4 & 12.8 & 73.4 & 55.65 \\
& EN-CoT          & 51.0  & 74.8  & 9.8  & 68.8  & 51.10  & 70.4 & 68.6 & 13.0 & 73.6 & 56.40 \\
& Average shift   & 58.6 & 77.4 & 4.8  & 74.2 & 53.75 & 65.4 & 67.0 & 9.4  & 70.0 & 52.95 \\
& All-step shift  & 72.4 & 87.6 & 11.4 & 83.2 & 63.65 & 88.8 & 89.0 & 55.8 & 88.0 & 80.40 \\
& Random-step     & 58.6 & 77.4 & 6.0  & 74.2 & 54.05 & 77.4 & 78.2 & 22.6 & 77.0 & 63.80 \\
& Critical-step   & 73.2 & 87.0 & 12.0 & 85.6 & 64.45 & 89.6 & 89.0 & 56.4 & 92.0 & 81.75 \\
\midrule

\multirow{6}{*}{TH}
& Baseline        & 53.0 & 78.6 & 10.6 & 70.6 & 53.20 & 68.6 & 68.8 & 14.8 & 71.8 & 56.00 \\
& EN-CoT          & 51.8  & 78.8  & 10.6  & 69.8  & 52.75 & 70.0 & 69.2 & 13.2   & 73.4 & 56.45 \\
& Average shift   & 59.0 & 78.8 & 5.2  & 75.0 & 54.50 & 66.4 & 67.4 & 9.2  & 67.4 & 52.60 \\
& All-step shift  & 76.0 & 86.4 & 11.0 & 82.6 & 64.00 & 86.8 & 88.6 & 59.6 & 92.0 & 81.75 \\
& Random-step     & 68.4 & 80.8 & 10.0 & 84.6 & 60.95 & 80.8 & 77.2 & 25.6 & 77.6 & 65.30 \\
& Critical-step   & 77.0 & 85.0 & 11.0 & 77.4 & 62.60 & 89.2 & 87.2 & 57.0 & 86.0 & 79.85 \\
\midrule

\multirow{6}{*}{PT}
& Baseline        & 62.6 & 96.6 & 12.0 & 83.0 & 63.55 & 75.4 & 75.6 & 13.4 & 73.0 & 59.35 \\
& EN-CoT          & 59.8  & 96.4  & 12.0  & 81.0  & 62.30  & 73.0 & 72.4 & 13.4 & 70.0 & 57.20 \\
& Average shift   & 64.0 & 97.0 & 11.8 & 84.0 & 64.20 & 72.4 & 73.6 & 8.6  & 67.6 & 55.55 \\
& All-step shift  & 83.4 & 98.4 & 11.0 & 86.0 & 69.70 & 90.0 & 89.0 & 58.2 & 84.0 & 80.30 \\
& Random-step     & 75.6 & 98.0 & 12.0 & 87.0 & 68.15 & 80.6 & 81.6 & 26.2 & 76.4 & 66.20 \\
& Critical-step   & 84.2 & 97.4 & 14.2 & 86.8 & 70.65 & 88.0 & 89.6 & 60.8 & 90.0 & 82.10 \\
\midrule

\multirow{6}{*}{VI}
& Baseline        & 59.2 & 91.8 & 10.8 & 76.4 & 59.55 & 69.6 & 72.6 & 13.6 & 73.2 & 57.25 \\
& EN-CoT          & 56.6  & 80.8  & 10.8  & 72.2  & 55.10  & 72.2 & 72.8 & 14.2   & 65.8 & 56.25 \\
& Average shift   & 59.0 & 89.4 & 4.8  & 75.0 & 57.05 & 69.0 & 71.8 & 10.2 & 65.8 & 54.20 \\
& All-step shift  & 76.0 & 92.6 & 11.0 & 86.6 & 66.55 & 89.4 & 89.0 & 56.2 & 92.0 & 81.65 \\
& Random-step     & 68.4 & 91.0 & 10.0 & 80.6 & 62.50 & 78.8 & 81.2 & 24.8 & 75.5 & 65.08 \\
& Critical-step   & 77.0 & 93.4 & 11.0 & 90.6 & 68.00 & 89.4 & 91.6 & 60.6 & 86.0 & 81.90 \\

\bottomrule
\end{tabular*}
\end{table*}

\end{document}